# Phase-Interface Instance Segmentation as a Visual Sensor for Laboratory Process Monitoring

Mingyue Li[1#], Xin Yang[1#], Shilin Yan [2], Jinye Ran [2], Morui Zhu [3], Huanqing Peng [4], Wei Peng [4], Guanghua Zhang [5], Shuo Li [1], Hao Zhang [2,4*]

[1] School of Chemistry and Chemical Engineering, Chongqing University of Technology, Chongqing 400054, China

[2] School of Chemistry and Chemical Engineering, Southwest University, Chongqing 400715, China

[3] Department of computer science, University of North Texas, Denton 76207, Texas, United States

[4] Hangzhou Digitalsalt Technology Co., Ltd., Hangzhou, 310000, China

[5] School of Big Data Intelligent Diagnosis and Treatment Industry, Taiyuan University, Taiyuan 030002, China

**Abstract:** Reliable visual monitoring of chemical experiments remains challenging in transparent glassware, where weak phase boundaries and optical artifacts degrade conventional segmentation. We formulate laboratory phenomena as the time evolution of phase interfaces and introduce Chemical transparent Glasses dataset 2.0 (CTG 2.0), a vessel-aware benchmark with 3,668 images, 23 glassware categories, and five multiphase interface types for phase-interface instance segmentation. Building on YOLO11m-seg, we propose LGA-RCM-YOLO, which combines Local-Global Attention (LGA) for robust semantic representation and Rectangular Self-Calibration

---

[*] Corresponding author: Hao Zhang, Email: haozhang@swu.edu.cn
#These authors contributed equally.

Module (RCM) for boundary refinement of thin, elongated interfaces. On CTG 2.0, the proposed model achieves 84.4% *AP*@0.5 and 58.43% *AP*@0.5-0.95, improving over the YOLO11m baseline by 6.42 and 8.75 AP points, respectively, while maintaining near real-time inference (13.67 FPS, RTX 3060). An auxiliary color-attribute head further labels liquid instances as colored or colorless with 98.71% precision and 98.32% recall. Finally, we demonstrate continuous process monitoring in separatory-funnel phase separation and crystallization, showing that phase-interface instance segmentation can serve as a practical visual sensor for laboratory automation.

# 1 Introduction

Autonomous and semi-autonomous chemical laboratories increasingly depend on online sensing to support reliable execution, monitoring, and decision-making(Dai et al., 2024; Porwol et al., 2020; Tom et al., 2024). In such settings, prescribed set-points and final assay results are often insufficient to characterize the realized process trajectory, particularly when multiphase behavior drives outcomes and failure modes (Shields et al., 2021; Wang et al., 2025). Across gas-liquid (G/L), liquid-liquid (L/L), liquid-solid (L/S), gas-solid (G/S), and solid-solid (S/S) systems, the most decision-relevant experimental phenomena are frequently expressed through phase-interface evolution, including phase emergence or disappearance, interface displacement and stability, phase-fraction variation, entrainment/ emulsification, and solid formation during crystallization/ precipitation (Seifrid et al., 2022; Xiouras et al., 2022). Consequently, converting visual phenomena into structured process descriptors via computer vision provides a general and efficient route to improve process observability (El-Khawaldeh et al., 2024; Manee et al., 2019). However, robust visual monitoring in laboratory glassware remains difficult: transparent vessels introduce refraction and specular reflections, phase boundaries can be weak and elongated, and practical deployment must accommodate diverse vessel geometries and scene variability (Xie et al., 2020; Zhang et al., 2022).

Computer vision has been progressively adopted as a non-contact sensing modality in chemical laboratory environments, spanning both foundational perception and system-level monitoring (Maaß et al., 2012; Vicente et al., 2019). The Vector-

LabPics work established large-scale recognition of vessels and material phases in realistic lab scenes, providing an important dataset and baseline for laboratory perception (Eppel et al., 2020). Complementarily, El-Khawaldeh et al. demonstrated that vision can deliver actionable online signals across practical laboratory operations by tracking cues such as liquid levels, homogeneity, turbidity, solids/residue, and color, moving toward closed-loop use(El-Khawaldeh et al., 2024). Together with the broader "computer vision as an analytical sensor" (CVAS) perspective in analytical chemistry, these studies motivate cameras as scalable measurement channels(Barrington et al., 2025; Buurma and Bagley, 2023). Nevertheless, for deployable multiphase monitoring, two limitations remain prominent: typed phase interfaces are rarely treated as first-class objects for instance-level segmentation and rigorous evaluation under transparent glass conditions, and robustness to weak/blurred interfaces across diverse vessels and backgrounds remains a practical bottleneck for transfer to real workflows (Li et al., 2025; Tom et al., 2024).

In this work, the problem is formulated as vessel-aware phase-interface instance segmentation for laboratory scenes. The formulation follows a physically consistent hierarchical perception logic: transparent vessels are first recognized to define constrained regions-of-interest (ROIs), and phase interfaces together with associated phase regions are subsequently segmented within the ROIs across the multiphase combinations encountered in practice (G/L, L/L, L/S, G/S, and S/S). This design is motivated by the observation that accurate ROI localization suppresses background confounders and stabilizes downstream segmentation in transparent scenes, while interface-centric parsing enables direct extraction of quantitative process descriptors (e.g., interface height, phase-fraction proxies, stability indices, and appearance attributes) (程晗 et al., 2023). Three contributions are made in this work:

(1) CTG 2.0 is constructed as a dedicated benchmark for vessel-aware phase/interface instance segmentation in chemical laboratories, covering diverse transparent vessels, interface categories, and realistic scene variability.

(2) LGA-RCM-YOLO is developed as a real-time segmentation framework that enhances weak, elongated interface structures via local-global context aggregation and directional self-calibration, improving the accuracy-efficiency trade-off against strong baselines.

(3) A streaming monitoring system is built around LGA-RCM-YOLO to perform real-time inference and event logging from industrial video streams, producing time-stamped masks, keyframes, and interface descriptors that support continuous separation and crystallization monitoring and can be archived to electronic lab records for downstream decision support.

The remainder of this paper is organized as follows. Section 2 reviews related work on computer vision as an online sensing modality in automated chemistry, laboratory scene understanding for transparent vessels and materials, and vision-based characterization of multiphase dynamics. Section 3 presents the CTG 2.0 dataset, including the problem definition, data sources, and dataset statistics. Section 4 details the proposed LGA-RCM-YOLO framework, including the LGA and RCM modules, the auxiliary color-attribute recognition head, and the experimental setup with evaluation protocols. Section 5 reports the results and discussion, covering overall benchmark performance and efficiency, interface-wise analysis, vessel-conditioned generalization, optical-contrast effects, and continuous process monitoring case studies. Section 6 concludes the paper and outlines limitations and future directions toward integrated, vision-driven monitoring and optimization in automated laboratory workflows.

## 2 Related Work

## 2.1 Computer vision as an online sensing modality in automated chemistry

Recent progress in automated and self-driving laboratories has made online state observability a central requirement, and camera-based monitoring has emerged as a practical, non-contact sensing layer (Li et al., 2025; Wei et al., 2025). Prior studies demonstrate that vision can support real-time monitoring and, in some cases, closed-loop control by tracking operational cues such as liquid level, homogeneity, turbidity, solids or residue, and color(Sasaki et al., 2024; Yao et al., 2024). For example, El-Khawaldeh et al. introduced HeinSight 2.0 for automated monitoring and control of diverse workup operations using multiple visual outputs, providing rapid acquisition of process-relevant signals; however, the system is typically configured around fixed vessel setups, which limits adaptability across heterogeneous glassware and viewing conditions(El-Khawaldeh et al., 2024). In parallel, Yan et al. proposed *Kineticolor* as a video analysis platform to extract kinetics-related information from color evolution in a Pd-catalyzed Miyaura borylation case study, illustrating the value of color trajectories but also reflecting a common emphasis in computer vision in analytical chemistry on liquid-phase chromaticity (Yan et al., 2023). Relatedly, Fyfe et al. used imaging-based kinetic data, again relying on colorimetric reactions and *Kineticolor*, to qualitatively assess CFD models of stirred-tank reactors, showing how vision-derived time series can inform process understanding in specific reactor classes (Fyfe et al., 2024). Collectively, these works validate the role of vision as a process sensor, yet most are built around task- or chemistry-specific visual proxies, such as liquid level or color, and therefore do not provide a unified representation that transfers across the broader set of multiphase operations encountered in laboratory practice. This motivates interface-centric perception, where the evolution of phase boundaries provides a generalizable visual descriptor for monitoring across liquid-liquid separation, gas-liquid systems, crystallization, and related workflows.

## 2.2 Laboratory scene understanding: vessels, materials, and transparent objects

Dataset-driven lab perception has primarily focused on recognizing what objects and material states are present, providing a foundation for scalable scene understanding in chemistry settings. Vector-LabPics by Eppel et al. includes 2,187 annotated images spanning common laboratory vessels and material appearances such as liquids, solids, foam, suspensions, and powders, enabling broad semantic parsing of laboratory scenes; however, it offers limited fine-grained differentiation of transparent glassware, uses relatively simple backgrounds, and does not explicitly cast experimental phenomena as the evolution of phase interfaces (Eppel et al., 2020). To better address transparent glassware perception, Ge Jiantong et al. constructed CTG (1,548 images) with the primary task of instance segmentation of transparent laboratory instruments, strengthening vessel contour delineation under reflections and overlap (葛建统 et al., 2023). From a complementary perspective, Wu et al. introduced LCDTC (5,916 well-annotated images) and standardized liquid-handling perception around container detection and liquid-level estimation, which is effective for height readouts but does not generalize to multiphase interface delineation beyond the liquid level (Wu et al., 2023). Collectively, these datasets establish key capabilities for vessel and material recognition in laboratory imagery, yet they provide limited support for vessel-aware, interface-centric monitoring where thin, deformable multiphase boundaries must be segmented as instances to produce time-resolved process descriptors.

## 2.3 Vision-based characterization of multiphase dynamics and interface-centric representation

Vision has been used to quantify dynamic behaviors relevant to chemical operations, including mixing, kinetics analysis, dispersed-phase characterization (e.g., droplets and particles), and workup monitoring such as extraction and crystallization (Barrington et al., 2022; Buurma and Bagley, 2023). Many existing pipelines rely on intensity/color proxies, dispersed-entity statistics, or operation-specific feature engineering (Li et al., 2025; Reid, 2025; Wei et al., 2025). In contrast, across common multiphase systems (G/L, L/L, L/S, G/S, and occasionally S/S), the dominant observable manifestation of process evolution is the formation, displacement, deformation, and stabilization of phase interfaces (El-Khawaldeh et al., 2024; 程晗 et al., 2023). This observation motivates an interface-centric formulation in which phase interfaces are treated as first-class, typed perception targets, providing a transferable representation for vision-based monitoring across diverse laboratory multiphase scenarios.

## 3 CTG 2.0 Dataset

### 3.1 Problem definition

CTG 2.0 is designed to support vision-based monitoring in chemical laboratories by representing multiphase experimental phenomena through phase-interface evolution. In typical laboratory operations, multiphase state changes are most directly observed as the appearance, displacement, deformation, and stabilization of interfaces. Accordingly, CTG 2.0 is organized for vessel-aware phase-interface instance segmentation: transparent vessels are treated as the structural carriers that define physically meaningful ROIs, while phase interfaces are annotated as first-class instance targets. Auxiliary objects such as labels and stoppers are also included because they frequently occur in practical scenes and can otherwise introduce systematic confounding near vessel boundaries and internal phase structures. This formulation is intended to support extraction of process-relevant descriptors for time-resolved monitoring.

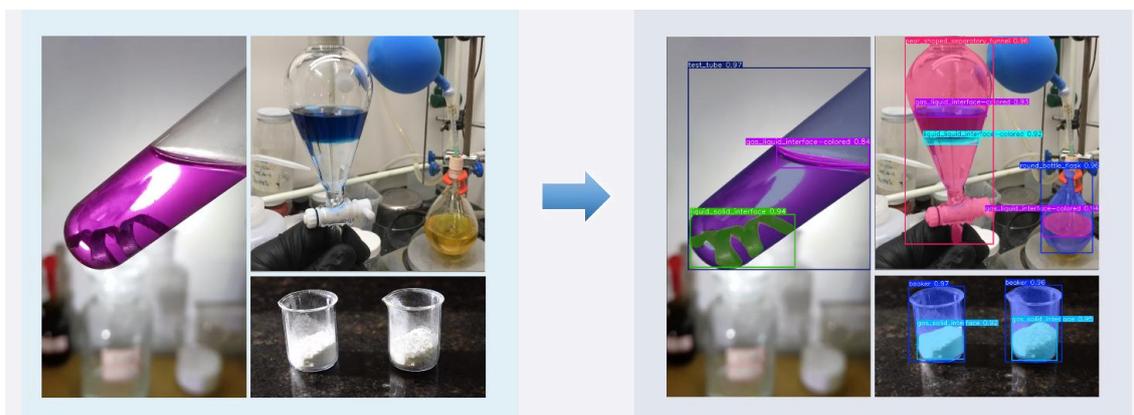

**Fig.1** Example of instance segmentation for glass containers and interfaces (including color recognition of liquid-liquid and gas-liquid interfaces).

## 3.2 Data sources and dataset overview

Images in CTG 2.0 were collected from three complementary sources to balance realism and diversity: (i) laboratory photos captured in chemical laboratories, (ii) selected samples from CTG 1.0, and (iii) frames extracted from publicly available online laboratory videos. The dataset contains 3,668 images, 30 categories, and 18,458 annotated instances. It covers substantial variation in illumination and background, as well as practical optical artifacts introduced by glass thickness, refraction, and specular reflections conditions that routinely occur in real laboratory deployment rather than controlled benchmark settings.

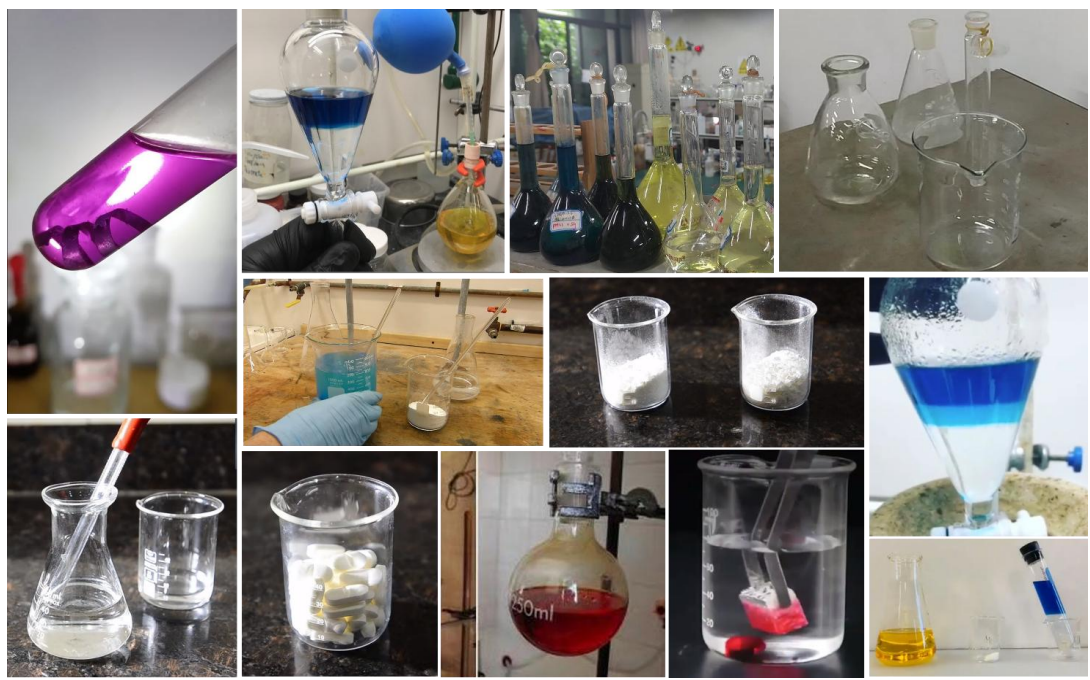

**Fig. 2**. samples of image in CTG 2.0 dataset.

## 3.3 Dataset statistics and splits

CTG 2.0 expands the original CTG dataset in both category coverage and scene complexity. Compared with CTG (1,548 images, 14 vessel types), CTG 2.0 includes 23 vessel categories and adds explicit multiphase interface annotations together with auxiliary objects, yielding 30 categories overall. Interface instances are naturally imbalanced, reflecting laboratory frequency and annotatability: 3,637 G/L, 852 L/S, 327 L/L, 477 G/S, and 7 S/S interface instances. The S/S class is extremely sparse because boundaries between solid phases are often visually indistinguishable and solid–solid contact is uncommon in typical laboratory imagery; therefore, this category is not statistically representative for learning-based evaluation and can be treated as a rare/auxiliary label in quantitative analysis. Scene density increases relative to CTG: the maximum number of instances per image rises from 80 to 112, with an average of approximately 5 instances per image. The instance-scale distribution spans small to large targets: 2,108 instances with area < 322 pixels (11.42%), 5,926 instances with area 322-962 pixels (32.11%), and 10,424 instances with area > 962 pixels (56.47%). For model development in this study, CTG 2.0 is split into 2,939 training images and 729 validation images.

## 4 Proposed Method

### 4.1 Baseline and notation

Given an image $I \in \mathbb{R}^{H \times W \times 3}$, instance segmentation aims to predict a set of instances $S = \{(c_i, M_i, b_i)\}_{i=1}^{N}$, where $c_i$ is the class label, $M_i$ is the instance mask, and $b_i$ is the bounding box. Our framework, LGA-RCM-YOLO, is built on YOLO11m-seg, which follows a backbone-neck-head design (Khanam and Hussain, 2024). YOLO11 introduces stronger feature extraction blocks (e.g., C3k2 as an enhanced variant of the

C2f-style CSP block) and places an attention refinement block (C2PSA) after multi-scale context aggregation (SPPF), improving representation capability under complex scenes (Chen et al., 2025).

## 4.2 Overall network structure

LGA-RCM-YOLO is built upon the YOLO11m-seg instance segmentation baseline. It preserves the original backbone, multi-scale neck, and segmentation head, and introduces two lightweight modifications designed for chemical reaction imagery dominated by transparent glassware, specular reflections, and weakly textured phase boundaries. The first modification augments high-level semantic extraction in the backbone through a Local-Global Attention (LGA) module (Shao, 2024). LGA is inserted immediately after the SPPF layer, where multi-scale context has already been aggregated, and before the subsequent attention-based refinement stage. This placement allows the network to refine high-level representations by simultaneously reinforcing local cues that are critical to interface perception, such as glass edges and meniscus transitions, and global dependencies that support continuity of long, thin interfaces across the vessel.

The second modification targets feature fusion in the neck through a C3k2_RCM block, where Rectangular Self-Calibration (RCM) is appended as a post-calibration unit rather than replacing the original C3k2 structure (Liu et al., 2024). In this design, C3k2 first performs CSP-style multi-branch feature transmission and bottleneck aggregation, followed by a channel unification step using a 1 × 1 convolution. RCM then operates on the fused feature maps and applies direction-sensitive calibration to amplify elongated, orientation-dependent structures while suppressing background responses induced by reflections or glass texture. This calibration is particularly beneficial for phase-interface masks whose contours are thin, deformable, and frequently aligned with vessel geometry.

The resulting multi-scale features are passed to the YOLO11 segmentation head to predict instance masks and interface categories, providing robust phase-interface segmentation under the optical artifacts and geometric constraints commonly encountered in laboratory and process monitoring.

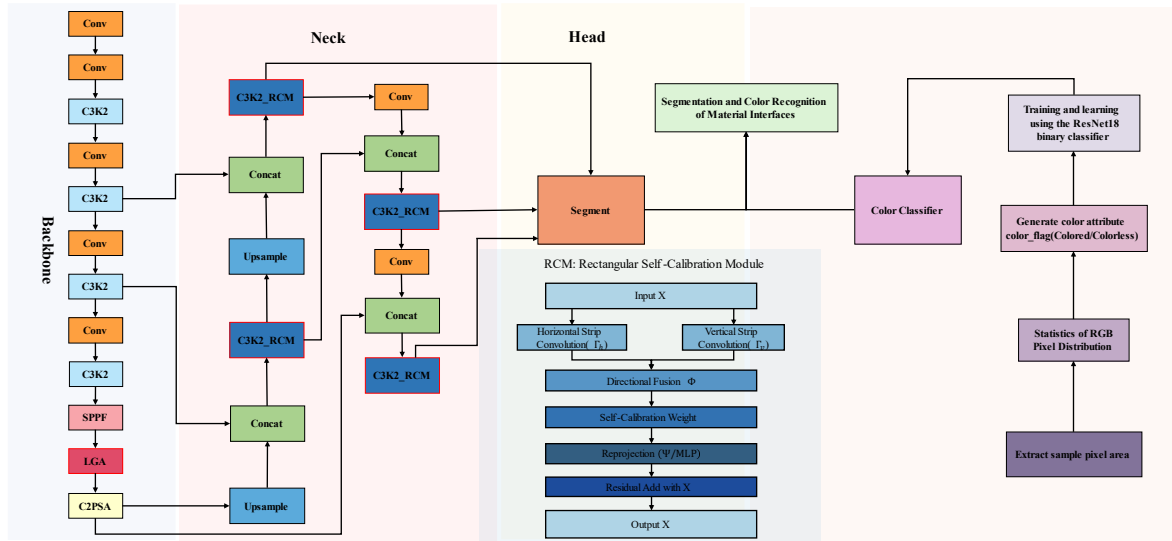

**Fig.3**. Overall architecture of the proposed LGA-RCM-YOLO framework for phase-interface instance segmentation in chemical reaction imagery.

## 4.3 Local-Global Attention module (LGA)

Transparent vessels and multiphase interfaces often exhibit weak texture and optical artifacts (refraction, reflection), requiring both fine local cues and global structural context. Let $X \in \mathbb{R}^{C \times h \times w}$ denote the input feature to LGA. LGA first builds a multi-scale representation in eq. (1):

$$X_{ms} = \sum_{s=1}^{S} \alpha_s f_s(X) \tag{1}$$

where $f_s(\cdot)$ denotes lightweight convolutions at different receptive fields and $\alpha_s$ are adaptive weights computed from pooled statistics.

LGA then aggregates local and global dependencies, which are shown in eq. (2) and (3):

$$X_{loc} = \sigma(A_{loc}(X_{ms})) \odot X_{ms} \tag{2}$$

$$X_{glob} = A_{glob}(X_{ms}) \tag{3}$$

where $A_{loc}(\cdot)$ denotes multi-kernel local attention and $A_{glob}(\cdot)$ denotes a global attention operator. The outputs are fused by a learnable gate:

$$\tilde{X} = \varphi(\gamma X_{loc} + (1-r) X_{glob}) \tag{4}$$

where $\gamma \in [0,1]$ and $\phi(\cdot)$ is a lightweight projection, and the resulting $\tilde{X}$ is passed to C2PSA and then to the neck.

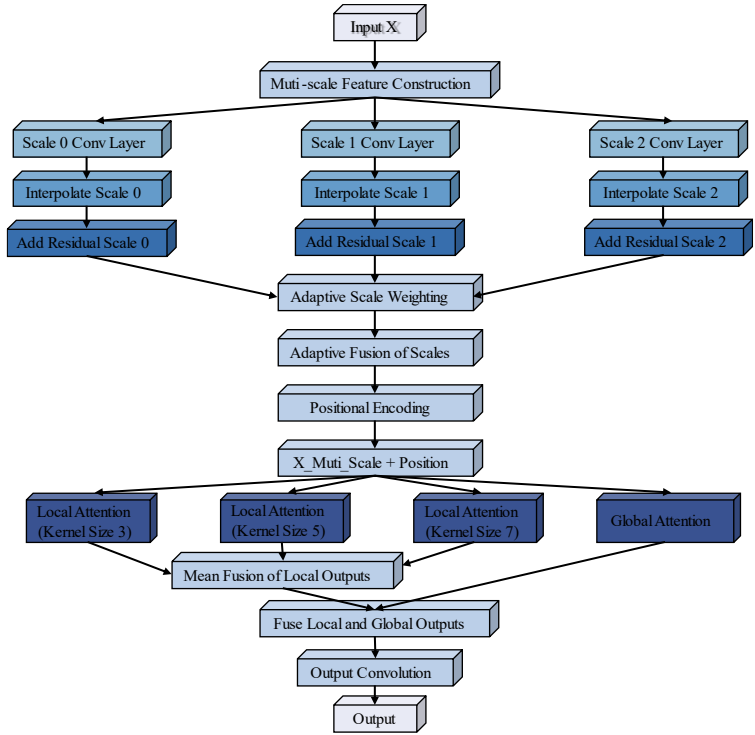

**Fig.4**. LGA module structure.

## 4.4 Rectangular Self-Calibration module (RCM) within C3k2

Phase interfaces frequently appear as thin, elongated structures with strong directional continuity. RCM introduces direction-constrained asymmetric modeling plus self-calibration to enhance structure-relevant responses while suppressing redundant background activations, under low computational overhead.

Let $X \in \mathbb{R}^{C \times h \times w}$ be the fused output feature of a C3k2 block. RCM extracts horizontal/vertical directional context:

$$X_h = R_h(X) \tag{5}$$

$$X_v = R_v(X) \tag{6}$$

where $R_h(\cdot)$ and $R_v(\cdot)$ are lightweight asymmetric operators.

Directional context is fused in eq. (7) with $[\cdot,\cdot]$ channel concatenation and $\Phi(\cdot)$ a lightweight fusion mapping.

$$D = \Phi([X_h, X_v]) \tag{7}$$

A self-calibration map is computed by eq. (8):

$$W = \sigma(G(D)) \tag{8}$$

where $G(\cdot)$ is a compact calibration mapping and $\sigma(\cdot)$ is Sigmoid.

A local detail branch is also computed in eq. (9).

$$L = \Psi(X) \tag{9}$$

Finally, calibrated features are reconstructed with a residual connection:

$$Y = P(W \odot L) + X \tag{10}$$

where $P(\cdot)$ is a projection for channel alignment. In the network, this RCM unit is appended after C3k2 fusion, forming C3k2_RCM in the neck.

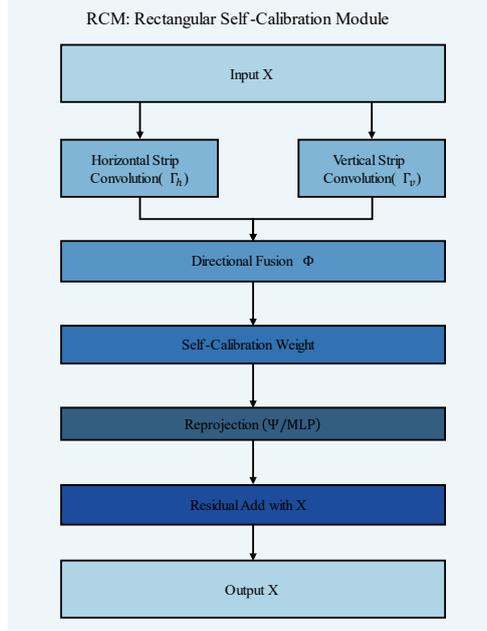

**Fig.5**. The principle of RCM

## 4.5 Auxiliary color-attribute recognition

To provide richer semantics for process interpretation, a weakly supervised binary color attribute is attached to liquid-related instances. Given an instance mask $M_i$, pixels within the mask region $\Omega_i=\{(x,y)|M_i(x,y)=1\}$ are used to compute simple RGB statistics to generate a pseudo-label $y_i^{color} \in [0, 1]$. Cropped instance regions are then used to train a ResNet-18 classifier for attribute prediction. The final output augments the segmentation result as class label and color property. This module does not modify the segmentation objective; it only appends an interpretable attribute to each predicted instance.

## 4.6 Experimental environment and Evaluation metrics

4.6.1 Experimental environment and training setup

All models were implemented in PyTorch (Python 3.9) on Windows 11 and trained and evaluated on an NVIDIA RTX 3060 GPU (12 GB). We reserve 729 images as a held-out test set and use the remaining 2,939 images for model development. Cross-validation is performed within the 2,939-image development split for hyperparameter selection and robustness checks; the final model is then trained using the full development split and evaluated once on the fixed 729-image test set. Unless stated otherwise, images are resized to 640 × 640, the batch size is 8, the optimizer is Adam, the initial learning rate is 0.001, and training runs for 200 epochs. Training and cross-validation curves, together with development-split statistics, are provided in the Supplementary Material. All performance numbers reported in the main text are computed on the held-out test set.

4.6.2 Evaluation protocol and metrics

We follow the COCO-style instance segmentation evaluation and report $AP@0.5$ and $AP@0.5\text{-}0.95$ to quantify segmentation quality under a single $IoU$ threshold and under the standard multi-threshold regime, respectively. In addition, we report Precision ($P$) and Recall ($R$) using an instance-matching rule at $IoU \geq 0.5$. For each image and class, predictions are matched one-to-one to ground-truth instances based on $IoU$, and a prediction is counted as a true positive if it matches an unmatched ground-truth instance of the same class; unmatched predictions are false positives and unmatched ground truths are false negatives. Precision and recall are computed from $TP$, $FP$, and $FN$ in the standard way. The same evaluation protocol is applied consistently across the overall benchmark, ablation studies, and interface-wise analyses. For vessel-conditioned analysis, interface correctness is evaluated under a hierarchical constraint that requires correct vessel recognition before interface matching; the conditioned protocol is defined explicitly in the corresponding results subsection.

## 5 Results and Discussion

### 5.1 Overall Performance, Efficiency, and Component Contribution

**Table 1** reports the benchmark results on CTG 2.0 for material phase-interface instance segmentation. LGA-RCM-YOLO attains the best overall performance and exceeds the strongest baseline, YOLO11m, by 6.42% in AP@0.5 and by 8.75% in AP@0.5-0.95. The larger gain under the stricter *IoU* regime is particularly important because it reflects improved boundary fidelity rather than only stronger coarse localization, which directly targets the dominant difficulty in transparent laboratory imagery where interfaces are weak-textured, reflective, and morphologically deformable. The advantage is consistent across YOLO variants and becomes more pronounced when compared with heavier instance-segmentation baselines that struggle in this setting; for example, AP@0.5-0.95 remains at 31.88% for ASPP-SOLOv2 and 45.80% for Mask R-CNN. The qualitative results in **Fig. 6** corroborate these trends, showing that competing methods more frequently miss instances or produce incomplete contours in multiphase scenes, whereas LGA-RCM-YOLO yields more continuous interface boundaries and clearer category assignment.

The accuracy improvements are obtained with modest computational overhead. Relative to YOLO11m, the computational cost increases from 123.7G to 135.2G Floating-Point Operations (FLOPs) (approximately 9.3%), while throughput decreases from 14.97 to 13.67 Frames Per Second (FPS) (approximately 8.7%), maintaining near-real-time feasibility for continuous monitoring. By comparison, ASPP-SOLOv2 is substantially heavier (210.53G FLOPs) and slower (4.15 FPS), and Mask R-CNN also operates at a lower speed (5.45 FPS) while exhibiting reduced robustness under interface ambiguity. Taken together, these results support the deployment-oriented positioning of LGA-RCM-YOLO as a practical perception module that improves boundary-consistent segmentation in reflective, transparent laboratory scenes without compromising operational efficiency.

Table.1 Instance segmentation comparison results (material phase interfaces).

| Models | P/% | R/% | mAP@0.5/% | mAP@0.5-0.95/% | FPS | FLOPs |
|---|---|---|---|---|---|---|
| Mask R-CNN | 45.80 | 55.50 | 76.18 | 45.80 | 5.45 | 75.01 |
| ASPP-SOLOv2 | 58.07 | 42.05 | 58.07 | 31.88 | 4.15 | 210.53 |
| YOLOv8m | 86.83 | 72.58 | 78.90 | 49.25 | 17.16 | 110.50 |
| YOLO11n | 73.23 | 71.20 | 73.05 | 43.50 | 16.43 | 10.40 |
| YOLO11m (baseline) | 86.68 | 71.03 | 77.98 | 49.68 | 14.97 | 123.70 |
| YOLO12m | 83.75 | 73.65 | 78.70 | 47.98 | 13.99 | 123.30 |
| OURS(LGA-RCM-YOLO) | 93.85 | 74.53 | 84.40 | 58.43 | 13.67 | 135.20 |

Table 2 reports the ablation results relative to the YOLO11m baseline, for which *AP*@0.5 is 77.98% and *AP*@0.5-0.95 is 49.68%. Introducing LGA raises *AP*@0.5 to 84.00% and *AP*@0.5-0.95 to 56.80%, corresponding to relative improvements of 7.72% and 14.33%, respectively. Introducing RCM alone increases *AP*@0.5 to 80.80% and *AP*@0.5-0.95 to 55.15%, which represents relative gains of 3.62% and 11.01%. The fact that RCM contributes more strongly to *AP*@0.5-0.95 than to *AP*@0.5 indicates that its primary effect lies in improving boundary fidelity under stricter overlap criteria, rather than in coarse interface localization. When both modules are enabled, the model achieves the best performance, reaching 84.40% on *AP*@0.5 and 58.43% on *AP*@0.5-0.95. These results correspond to relative improvements of 8.23% and 17.61% over the baseline, and the combined configuration remains superior to single-module variant on *AP*@0.5-0.95. Overall, the ablation pattern supports a complementary interaction in which LGA stabilizes high-level semantics under optical artifacts to reduce interface-region ambiguity, while RCM strengthens contour continuity during neck fusion for thin and elongated interfaces, thereby yielding the most consistent improvements under strict *IoU* evaluation.

**Table 2** Ablation study on individual module modifications and combinations (material phase interfaces).

| Experiment | LGA | RCM | *P*/% | *R*/% | mAP@0.5/% | mAP@0.5-0.95/% |
|---|---|---|---|---|---|---|
| 1 | | | 86.98 | 71.03 | 77.98 | 49.68 |
| 2 | √ | | 91.38 | 75.40 | 84.00 | 56.80 |
| 3 | | √ | 90.48 | 72.25 | 80.80 | 55.15 |
| 4 | √ | √ | 93.85 | 74.53 | 84.40 | 58.43 |

In addition to segmentation, the auxiliary color-attribute head achieves strong classification performance, with a precision of 98.71% and a recall of 98.32%. The predicted color attribute is attached to liquid-related instances as a "class plus color property" descriptor, which enriches downstream semantic interpretation of reaction and separation states as illustrated in **Fig. 6**, while leaving the primary segmentation objective unchanged.

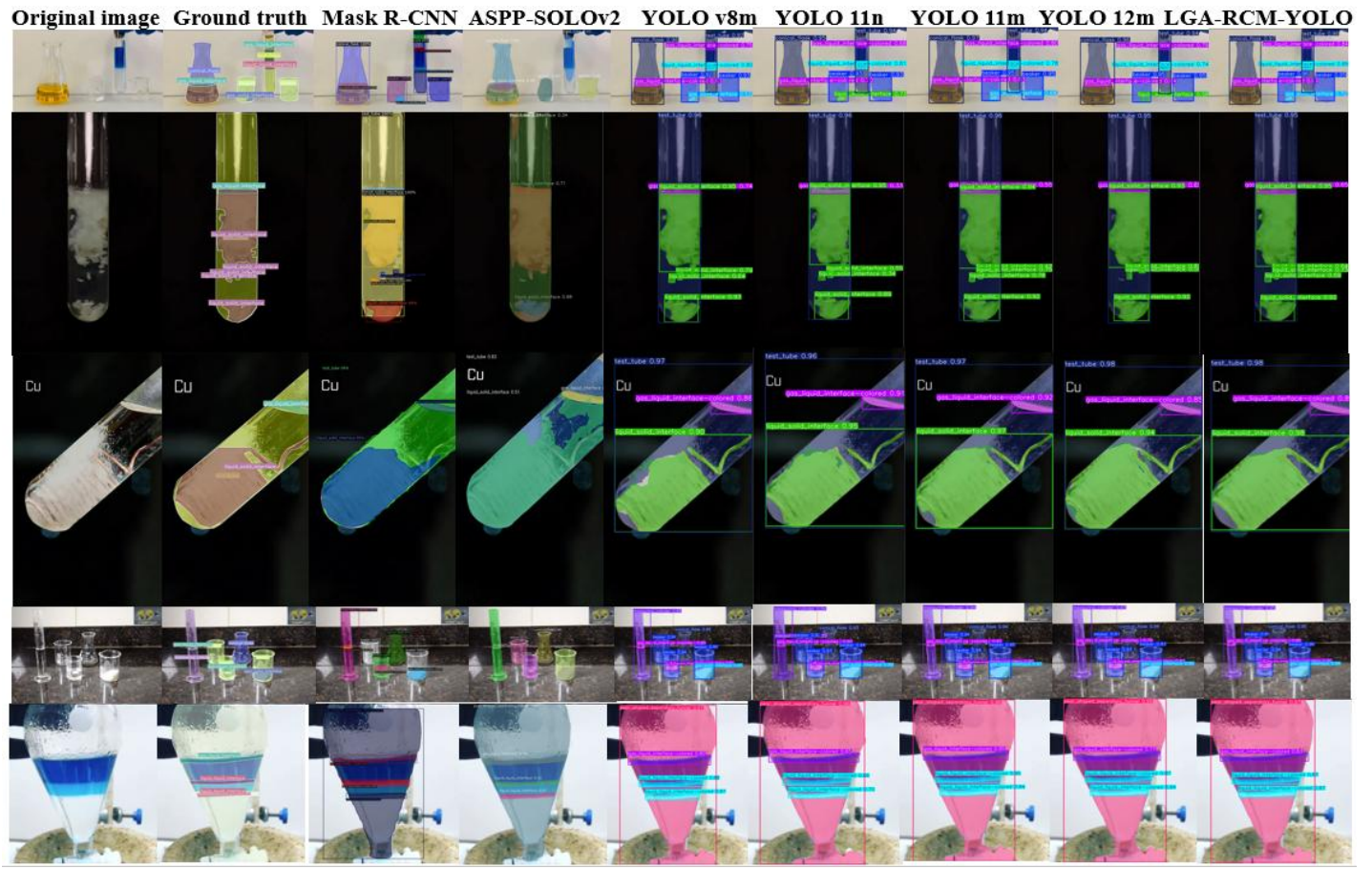

**Fig.6**. Visualization comparison of instance segmentation results.

## 5.2 Interface-wise performance across multiphase categories

Across interface types, LGA-RCM-YOLO exhibits a chemically and optically consistent performance profile (**Table 3**). Under the boundary-sensitive metric *AP*@0.5-0.95, G/S is the most reliable category (67.33%), followed by L/S (58.39%) and L/L (56.87%), whereas G/L remains the most challenging (51.12%) despite achieving an AP@0.5 of 78.64%. The gap between *AP*@0.5 and *AP*@0.5-0.95 for G/L reaches 27.52%, indicating that the interface region is generally detected, whereas accurate contour delineation is impaired by specular highlights, meniscus curvature, and refraction in transparent vessels; these optical effects attenuate true boundaries and introduce competing gradient structures. Liquid-involved interfaces (L/L, L/S) benefit from stronger geometric constraints, such as stratification boundaries and wall-contact structure, which supports higher strict-*IoU* quality; however, they remain sensitive to weak contrast and wetting/adhesion near glass walls, as reflected by the persistent divergence between AP@0.5 and AP@0.5-0.95. For L/L, AP@0.5 is 89.77%, whereas AP@0.5-0.95 decreases to 56.87%, corresponding to a 32.9 percentage-point gap. Given the class imbalance in **Table 3** and the limited support for S/S, these trends are interpreted as observability-driven and are examined further through vessel-conditioned generalization and low-contrast L/L analysis in the following sections.

**Table 3**. Per-interface instance segmentation performance of LGA-RCM-YOLO on CTG 2.0 (phase-interface categories). Macro-averaged scores are computed as unweighted averages over interface categories with sufficient support with the S/S excluded.

| interface type | instance number | P/% | R/% | mAP@0.5/% | mAP@0.5-0.95/% | Δ(AP0.5-AP0.5-0.95)/% |
|---|---|---|---|---|---|---|
| G/L | 1079 | 93.50 | 66.67 | 78.64 | 51.12 | 27.52 |
| G/S | 146 | 91.82 | 80.0 | 85.02 | 67.33 | 17.69 |
| L/L | 95 | 96.59 | 79.69 | 89.77 | 56.87 | 32.90 |
| L/S | 261 | 92.94 | 69.54 | 83.94 | 58.39 | 25.55 |
| S/S | 3 | / | / | / | / | |
| Macro | / | 93.71 | 73.98 | 84.34 | 58.43 | 25.92 |

## 5.3 Vessel-conditioned analysis and generalization across glassware

**Table 4** reports vessel-conditioned performance under a hierarchical matching rule, where an interface prediction is counted as a true positive only after correct vessel recognition (Level 1) and correct within-vessel interface matching (Level 2). Because several vessel-interface strata contain limited instances, particularly those with fewer than 20 samples, these rows are interpreted as indicative; the discussion therefore emphasizes trends supported by higher-count settings. Under this deployment-oriented criterion, gas–liquid performance is strongly vessel dependent: round-bottom flasks and conical flasks achieve high strict-*IoU* quality (*AP*@0.5-0.95 of 71.49% with $n = 50$, and 65.20% with $n = 73$), whereas the volumetric flask constitutes a consistent failure mode with the lowest *AP*@0.5-0.95 of 26.58% despite the largest support ($n = 204$). This contrast is consistent with vessel optics and geometry, since narrow neck/shoulder regions concentrate specular reflections and compress the visible interface into fewer pixels, which disproportionately penalizes boundary fidelity. For L/L and L/S interfaces, **Table 4** shows a recurring pattern in which recall remains high in several settings while precision is substantially lower, indicating that the interface region is often detected but mask quality is degraded by fragmentations and boundary leakage when contrast is weak, interfaces deform, or wetting effects merge liquid evidence with glass or solid texture. These container-induced failure modes are expected to intensify in curved or constricted vessels and motivate the subsequent hard-case analyses that explicitly disentangle generalization across glassware from low-contrast interface observability.

**Table 4** Container-Conditioned Instance Segmentation Performance of Different phase interface.

| interface | vessel type | instance number | P/% | R/% | mAP@0.5/% | mAP@0.5-0.95/% |
|---|---|---|---|---|---|---|
| G/L | beaker | 130 | 44.68 | 81.82 | 79.12 | 59.16 |
| | conical flask | 73 | 46.21 | 95.31 | 91.28 | 65.20 |
| | pear-shaped separatory funnel | 37 | 53.57 | 96.77 | 95.17 | 49.74 |
| | round-bottom flask | 50 | 64.79 | 97.87 | 96.54 | 71.49 |
| | test tube | 86 | 45.65 | 95.45 | 91.67 | 49.29 |
| | volumetric flask | 204 | 24.53 | 63.73 | 57.34 | 26.58 |
| L/S | beaker | 39 | 23.10 | 99.99 | 99.99 | 88.31 |
| | conical flask | 9 | 20.69 | 99.99 | 99.99 | 84.58 |
| | pear-shaped | 3 | 22.22 | 66.67 | 66.67 | 39.99 |
|     | separatory funnel         |    |       |       |       |       |
|     | round-bottom flask        | 8  | 41.67 | 83.33 | 83.33 | 64.99 |
|     | test tube                 | 83 | 33.33 | 74.99 | 74.99 | 50.83 |
| G/S | beaker                    | 41 | 45.10 | 91.99 | 91.99 | 72.45 |
|     | conical flask             | 7  | 46.15 | 99.99 | 99.99 | 47.01 |
|     | round-bottom flask        | 6  | 49.99 | 99.99 | 97.62 | 96.19 |
|     | test tube                 | 16 | 58.33 | 87.50 | 87.50 | 76.04 |
| L/L | pear-shaped separatory funnel | 19 | 38.10 | 99.99 | 96.96 | 54.10 |
|     | test tube                 | 23 | 38.46 | 99.99 | 99.99 | 81.99 |

## 5.4 Effect of Optical Contrast on L/L Interface Segmentation

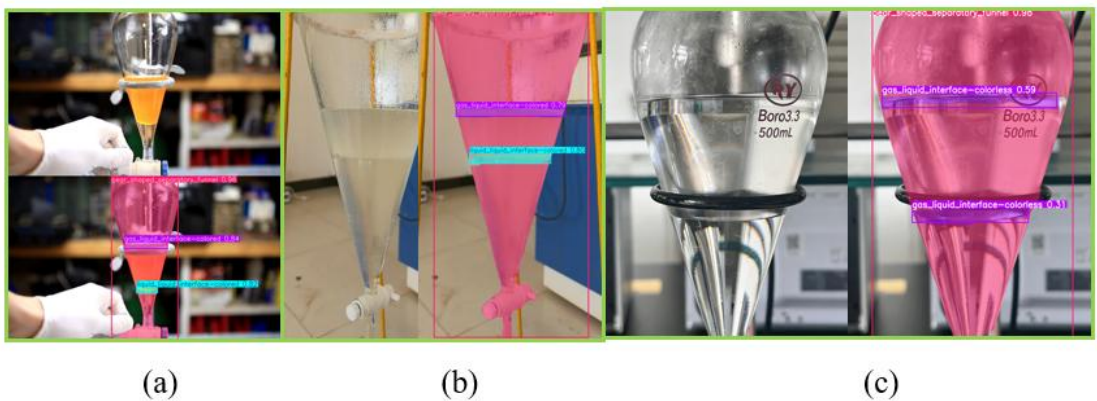

(a)　　　　　　　　(b)　　　　　　　　(c)

**Fig. 7**. Representative liquid–liquid (L/L) interface scenarios in CTG 2.0 illustrating the impact of optical contrast on interface observability and segmentation difficulty: (a) L/L interface with a large chromatic difference between phases; (b) L/L interface with similar but slightly translucent colors; (c) fully transparent L/L interface with a physical separation boundary, where visibility relies primarily on subtle refraction/specular effects and segmentation becomes most challenging.

It is highlighted in **Fig. 7** that L/L segmentation accuracy is fundamentally governed by interface observability, which in practice is dominated by the optical contrast budget between the two phases. When the two liquids exhibit a large chromatic difference, the interface produces a stable, high-SNR transition in both color and intensity; under such conditions the network is effectively solving a well-posed boundary-localization problem and therefore yields confident, geometrically complete masks. When colors are similar, but the phases remain slightly translucent, the interface becomes a low-contrast boundary whose visibility is carried mainly by weak cues such as subtle shading gradients, meniscus curvature, and local specular patterns shaped by illumination and vessel curvature. So, the model typically still localizes the interface but with softer boundaries and reduced confidence, reflecting sensitivity to lighting and background reflections rather than a failure to understand the phase change. The fully transparent case is qualitatively different: if chromatic and luminance contrasts vanish, the remaining information comes largely from refractive-index discontinuities (weak Fresnel reflections, small geometric distortions, and occasional highlight shifts), which are unstable in single-view RGB images and can be dominated by glass glare, lens exposure, and small perturbations in camera pose; in other words, the task becomes close to ill-posed without controlled illumination, multi-view geometry, or temporal cues. This explains why the model is prone to miss detections or fragmentary masks in **Fig. 7(c)** even if it performs well on **Fig. 7(a-b)**, and it also clarifies that the weakness is driven by a combination of data scarcity (such scenes are rare and hard to record with consistent quality) and intrinsic visual ambiguity. From a chemical-process perspective, this is not merely a modeling issue: reliable monitoring of fully transparent L/L separation typically requires either engineered observability (e.g., backlighting/ polarization, structured illumination, multi-view, or time-consistent tracking) or complementary sensing modalities, and this category therefore represents the most meaningful bottleneck for general-purpose vision-based reaction monitoring.

**5.5 Continuous process monitoring case studies**

5.5.1 System Implementation for Streaming Inference and Event Logging

To support continuous, decision-relevant monitoring in laboratory conditions, we implement an integrated hardware-software pipeline that turns phase-interface segmentation into a real-time visual sensor suitable for fume-hood operation. A 1080p industrial camera is mounted on an adjustable stand and connected via Ethernet to a laptop placed inside the hood; the camera pose is set to face the target vessel directly, and moderate off-axis viewing (within 15° for both elevation and depression angles) remains acceptable, enabling stable tracking of one or multiple vessels within the field of view. The live stream is pulled from the camera's native RTSP endpoint (compatible with H.264/ H.265), re-packaged and optimized into frame data using FFmpeg, and pushed via RTMP to a Nginx relay to ensure robust transport and low-latency visualization on an external display. On the processing side, LGA-RCM-YOLO is deployed in PyTorch and performs online frame decoding, vessel/interface inference, and instance annotation; critically, the segmentation masks are immediately converted into process descriptors within predefined statistical regions (e.g., interface height, area, or inter-interface distance), so the user interface presents not only overlays but also quantitative trends that are interpretable for operational decisions. System states and outputs are synchronized through a Redis real-time database: event triggers emit start signals, end conditions close the loop, and the most confident key frames are archived as structured evidence, enabling reproducible traceability. Finally, the video-derived descriptors and key frames can be synchronized over wireless networking to an electronic lab notebook, forming a multimodal experiment record that complements conventional logs (temperature, stirring, pH, spectra) and supports downstream analysis for process optimization and automation.

5.5.2 Liquid-liquid separation process in the separatory funnel

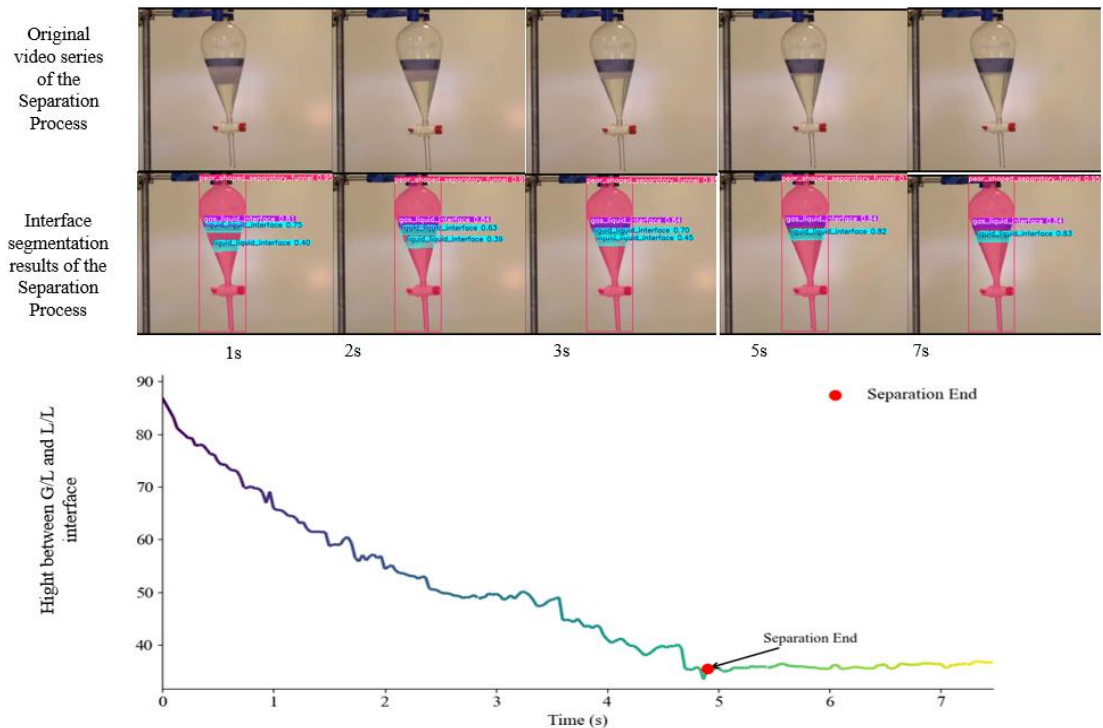

**Fig. 8**. Continuous monitoring of a liquid–liquid separation process in a pear-shaped separatory funnel using phase-interface instance segmentation. Top row: representative frames from the original video sequence. Middle row: corresponding interface segmentation results (vessel ROI with detected G/L and L/L interfaces). Bottom: time series of the vertical distance $\mathit{\Delta}h(t)$ between the G/L and L/L interfaces.

In the separatory-funnel case (**Fig. 8**), the practical challenge is that emulsified droplets prolong phase disengagement and make endpoint judgement subjective when relying on visual inspection alone; to stress-test continuous tracking, the original long process is temporally compressed into a 7s clip while preserving the characteristic dynamics. During separation, the system simultaneously tracks the G/L and L/L interfaces and converts the segmentation masks into a physically interpretable descriptor, the vertical distance $\Delta h(t)$ between the two interfaces. This choice is chemically meaningful: as dispersed droplets coalesce and the intermediate emulsion collapses, the bulk L/L boundary progressively approaches the G/L surface and then becomes quasi-stationary once the two liquid phases fully stratify, so $\Delta h(t)$ should decrease and then plateau. The extracted trajectory follows this expected mechanism: an initially rapid decline driven by droplet motion and interfacial perturbations, followed by a clear stabilization around 5s where the slope approaches zero, allowing the separation endpoint to be detected by a simple stationarity criterion. The automatically marked endpoint (red marker) aligns well with human judgement, while providing stronger interpretability than image-entropy heuristics because it measures a direct geometric consequence of phase disengagement rather than an indirect proxy sensitive to illumination and background variation.

5.5.3 crystallization monitoring by solid area evolution

In the crystallization case study (**Fig. 9**), phase-interface segmentation is used to convert visual observations of a supersaturated sodium acetate solution into a quantitative kinetic descriptor. Prior to seeding, the solution remains in a metastable supersaturated state and the predicted solid mask is essentially absent, consistent with no nucleation. After the seed is introduced (around 1-2 s), a non-zero solid region is detected, marking the onset of crystallization; subsequent growth is reflected by a sustained increase in solid area, which provides a direct, interpretable proxy for crystal mass/volume evolution under fixed imaging geometry. The short-term fluctuations and occasional local decreases in the area trace are expected in practice, arising from crystal motion, partial occlusion, surface reflections, and boundary ambiguity at early stages when crystals are sparse and translucent; however, the overall monotonic upward trend indicates entry into a stable growth regime. The abrupt jump near 6.6 s is attributed to video editing rather than a true burst nucleation event, underscoring an important deployment point: time-series descriptors derived from segmentation should be interpreted together with acquisition metadata, and can be readily stabilized using simple temporal smoothing or change-point logic when required for closed-loop control.

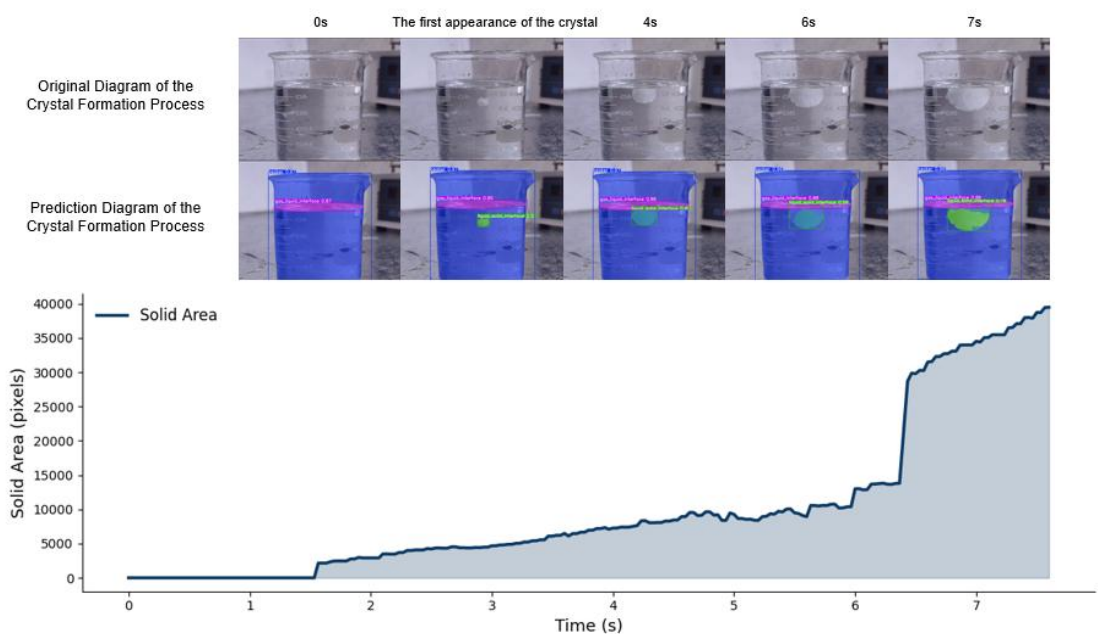

**Fig. 9**. Continuous crystallization monitoring via solid-phase segmentation in a supersaturated sodium acetate experiment. Top: representative frames from the original video and corresponding model predictions. Bottom: time series of predicted solid area (pixels), where the first non-zero area indicates nucleation after seeding and the subsequent rise reflects crystal growth.

# 6 Conclusion

This study advances computer vision sensing for chemical laboratories by treating experimental phenomena as the time evolution of phase interfaces and formulating the core perception task as vessel-aware phase-interface instance segmentation. To support research and reproducible comparison in transparent, reflection-dominated laboratory scenes, we release CTG 2.0, a curated benchmark that covers diverse glassware categories and multiphase interface types encountered in real experiments. Building on a strong one-stage baseline, we develop LGA-RCM-YOLO by introducing Local–Global Attention to strengthen high-level semantic representations and a Rectangular Self-Calibration unit to improve direction-sensitive boundary refinement. Extensive experiments on CTG 2.0 demonstrate consistent accuracy gains, particularly under strict *IoU* evaluation that better reflects phase-boundary fidelity, while preserving near real-time inference. In addition, an auxiliary color-attribute head provides reliable "class plus color property" descriptors for liquid-related instances, enriching semantic interpretation for downstream analysis. Finally, system-level demonstrations in liquid-liquid separation and crystallization monitoring show that segmentation-derived interface descriptors can be used to quantify process dynamics and support endpoint and state assessment.

Future work will focus on improving robustness for low-contrast interfaces and geometry-challenging glassware by combining targeted data collection, optics-aware augmentation, and temporally consistent inference. We will extend interface descriptors toward richer process measures, such as droplet-size evolution and crystallization kinetics, and integrate vision with complementary sensors for more reliable closed-loop monitoring.

# Declaration of competing interest

The authors declare that they have no known competing financial interests or personal relationships that could have appeared to influence the work reported in this paper.

# Acknowledgment

This study was supported by Hangzhou Digitalsalt Technology Research Fund (HDTF0001), National Natural Science Foundation of China (21806131), Chongqing Talent Project (No. CQYC202203091214) and .

# References

Barrington, H., Dickinson, A., McGuire, J., Yan, C., Reid, M., 2022. Computer vision for kinetic analysis of lab-and process-scale mixing phenomena. Organic Process Research & Development 26, 3073-3088.

Barrington, H., McCabe, T., Donnachie, K., Fyfe, C., McFall, A., Gladkikh, M., McGuire, J., Yan, C., Reid, M., 2025. Parallel and high throughput reaction monitoring with computer vision. Angewandte Chemie International Edition 64, e202413395.

Buurma, N.J., Bagley, S.W., 2023. A focus on computer vision for non-contact monitoring of catalyst degradation and product formation kinetics. Chemical Science 14, 10994-10996.

Chen, X., Jiang, N., Yu, Z., Qian, W., Huang, T., 2025. Citrus leaf disease detection based on improved YOLO11 with C3K2, International Conference on Computer Graphics, Artificial Intelligence, and Data Processing (ICCAID 2024). SPIE, pp. 746-751.

Dai, T., Vijayakrishnan, S., Szczypiński, F.T., Ayme, J.-F., Simaei, E., Fellowes, T., Clowes, R., Kotopanov, L., Shields, C.E., Zhou, Z., 2024. Autonomous mobile robots for exploratory synthetic chemistry. Nature 635, 890-897.

El-Khawaldeh, R., Guy, M., Bork, F., Taherimakhsousi, N., Jones, K.N., Hawkins, J.M., Han, L., Pritchard, R.P., Cole, B.A., Monfette, S., 2024. Keeping an "eye" on the experiment: computer vision for real-time monitoring and control. Chemical Science 15, 1271-1282.

Eppel, S., Xu, H., Bismuth, M., Aspuru-Guzik, A., 2020. Computer vision for recognition of materials and vessels in chemistry lab settings and the vector-labpics data set. ACS central science 6, 1743-1752.

Fyfe, C., Barrington, H., Gordon, C.M., Reid, M., 2024. A Computer Vision Approach toward Verifying CFD Models of Stirred Tank Reactors. Organic Process Research & Development 28, 3661-3673.

Khanam, R., Hussain, M., 2024. Yolov11: An overview of the key architectural enhancements. arXiv preprint arXiv:2410.17725.

Li, Y., Dutta, B., Yeow, Q.J., Clowes, R., Boott, C.E., Cooper, A.I., 2025. High-throughput robotic colourimetric titrations using computer vision. Digital Discovery 4, 1276-1283.

Liu, Y., Wang, Y., Wu, C., Wan, G., Yashchyshyn, Y., 2024. Self-Calibration Technique and Permittivity Measurement Using a Slotted Rectangular Waveguide. IEEE Transactions on Microwave Theory and Techniques.

Maaß, S., Rojahn, J., Hänsch, R., Kraume, M., 2012. Automated drop detection using image analysis for online particle size monitoring in multiphase systems. Computers & Chemical Engineering 45, 27-37.

Manee, V., Zhu, W., Romagnoli, J.A., 2019. A Deep Learning Image-Based Sensor for Real-Time Crystal Size Distribution Characterization. Industrial & Engineering Chemistry Research 58, 23175-23186.

Porwol, L., Kowalski, D.J., Henson, A., Long, D.L., Bell, N.L., Cronin, L., 2020. An autonomous chemical robot discovers the rules of inorganic coordination chemistry without prior knowledge. Angewandte Chemie 132, 11352-11357.

Reid, M., 2025. Computer vision for mixing analysis: Transforming cameras into quantitative tools for chemical process understanding, Advances in Physical Organic Chemistry. Elsevier, pp. 65-119.

Sasaki, R., Fujinami, M., Nakai, H., 2024. Application of object detection and action recognition toward automated recognition of chemical experiments. Digital Discovery 3, 2458-2464.

Seifrid, M., Pollice, R., Aguilar-Granda, A., Morgan Chan, Z., Hotta, K., Ser, C.T., Vestfrid, J., Wu, T.C., Aspuru-Guzik, A., 2022. Autonomous chemical experiments: Challenges and perspectives on establishing a self-driving lab. Accounts of Chemical Research 55, 2454-2466.

Shao, Y., 2024. Local-Global Attention: An Adaptive Mechanism for Multi-Scale Feature Integration. arXiv preprint arXiv:2411.09604.

Shields, B.J., Stevens, J., Li, J., Parasram, M., Damani, F., Alvarado, J.I.M., Janey, J.M., Adams, R.P., Doyle, A.G., 2021. Bayesian reaction optimization as a tool for chemical synthesis. Nature 590, 89-96.

Tom, G., Schmid, S.P., Baird, S.G., Cao, Y., Darvish, K., Hao, H., Lo, S., Pablo-García, S., Rajaonson, E.M., Skreta, M., 2024. Self-driving laboratories for chemistry and materials science. Chemical Reviews 124, 9633-9732.

Vicente, A., Raveendran, R., Huang, B., Sedghi, S., Narang, A., Jiang, H., Mitchell, W., 2019. Computer vision system for froth-middlings interface level detection in the primary separation vessels. Computers & Chemical Engineering 123, 357-370.

Wang, M., Jia, Z., Peng, H., Peng, W., Ran, J., Zhu, J., Zhao, L., Zhai, C., Yang, X., Zhang, H., 2025. Green Chemistry-Oriented Multiobjective Optimization with Semiautomated Process Monitoring for Apixaban Intermediate Synthesis. ACS Sustainable Chemistry & Engineering 13, 12134-12145.

Wei, H.-L., Ma, X.-M., Qin, J.-Z., Su, Y.-H., Luo, Z.-H., 2025. Image-based method for in-situ monitoring of reaction kinetics. Chemical Engineering Science, 122720.

Wu, Y., Ye, H., Yang, Y., Wang, Z., Li, S., 2023. Liquid Content Detection In Transparent Containers: A Benchmark. Sensors 23, 6656.

Xie, E., Wang, W., Wang, W., Ding, M., Shen, C., Luo, P., 2020. Segmenting transparent objects in the wild, European conference on computer vision. Springer, pp. 696-711.

Xiouras, C., Cameli, F., Quilló, G.L., Kavousanakis, M.E., Vlachos, D.G., Stefanidis, G.D., 2022. Applications of Artificial Intelligence and Machine Learning Algorithms to Crystallization. Chemical Reviews 122, 13006-13042.

Yan, C., Cowie, M., Howcutt, C., Wheelhouse, K.M., Hodnett, N.S., Kollie, M., Gildea, M.,

Goodfellow, M.H., Reid, M., 2023. Computer vision for non-contact monitoring of catalyst degradation and product formation kinetics. Chemical Science 14, 5323-5331.

Yao, T., Liu, J., Wan, X., Li, B., Rohani, S., Gao, Z., Gong, J., 2024. Deep-learning based in situ image monitoring crystal polymorph and size distribution: Modeling and validation. AIChE Journal 70, e18279.

Zhang, J., Yang, K., Constantinescu, A., Peng, K., Müller, K., Stiefelhagen, R., 2022. Trans4Trans: Efficient transformer for transparent object and semantic scene segmentation in real-world navigation assistance. IEEE Transactions on Intelligent Transportation Systems 23, 19173-19186.

程晗, 祝模芮, 孔新淋, 彭焕庆, 彭伟, 张浩, 2023. 基于边缘检测的分液过程监测与终点识别. 中国安全科学学报 33, 107.

葛建统, 杨鑫, 祝模芮, 冉进业, 翟持, 张浩, 2023. 基于 ASPP-SOLOv2 的复杂场景下透明玻璃仪器实例分割. 高校化学工程学报 37, 962-970.